%% file: main.tex
\documentclass[conference]{IEEEtran}
\IEEEoverridecommandlockouts

\PassOptionsToPackage{bookmarks=false,hidelinks}{hyperref}
\usepackage[utf8]{inputenc}
\usepackage[T1]{fontenc}
\usepackage{amsmath,amssymb,amsfonts}
\usepackage{graphicx}
\usepackage{textcomp}
\usepackage{xcolor}
\definecolor{optblue}{RGB}{0,114,178}
\definecolor{newverm}{RGB}{213,94,0}
\usepackage{tikz}
\usetikzlibrary{arrows.meta,positioning,fit,backgrounds,calc}
\usepackage{pgfplots}
\pgfplotsset{compat=1.18}
\usepackage{array}
\usepackage{booktabs}
\usepackage{eso-pic}  %
\usepackage{algorithm}
\usepackage[noend]{algpseudocode}
\usepackage[acronym,nomain,nonumberlist,nopostdot,nogroupskip]{glossaries}
\usepackage{orcidlink}
\usepackage[%
    backend=biber, hyperref=false, sorting=none, natbib=true,
    language=english, defernumbers=true, maxbibnames=2, minbibnames=1,
    pagetracker=false, citecounter=false, citetracker=true, style=ieee,
]{biblatex}
\renewbibmacro*{volume+number}{%
  \printfield{volume}%
  \iffieldnums{number}{\mkbibparens{\printfield{number}}}{}%
}

\tikzset{
  blk/.style={draw, rounded corners=1pt, align=center, font=\scriptsize,
              inner sep=3pt, minimum height=1.6em},
  flow/.style={-{Stealth[length=2mm]}, thick},
  flowback/.style={-{Stealth[length=2mm]}, thick, dashed},
  lbl/.style={font=\tiny, midway, fill=white, inner sep=1pt},
}

\newif\ifdraftmarks\draftmarksfalse

\newif\ifackfunding\ackfundingtrue
\newif\ifaidisclosure\aidisclosuretrue

\newif\ifpreprint\preprinttrue

\newif\iftierii\tieriifalse
\newcommand{\tii}[1]{\iftierii#1\fi}

\newcommand{\openitems}{\ifdraftmarks
\begin{center}\footnotesize\color{red!75!black}
\fbox{\parbox{0.95\columnwidth}{\raggedright\footnotesize
\textbf{DRAFT --- all open items resolved.} R7 closed 2026-08-19 by the
pre-registered three-arm lmmse-pair ablation (cost-null, artifact-quality
effect in the no-documentation arm; Secs.~I, V, VI;
record: \texttt{analysis/ablation/}). Set
\texttt{\textbackslash draftmarksfalse} for the submission build.\\
}}\end{center}\fi}

\begin{document}
\ifpreprint
\AddToShipoutPictureBG*{%
  \AtPageUpperLeft{\put(\LenToUnit{\dimexpr(\paperwidth-\textwidth)/2\relax},\LenToUnit{-0.40in}){%
    \parbox[t]{\textwidth}{\centering\footnotesize
      \textbf{--- PREPRINT ---}\\
      Accepted version of a paper accepted at the 2026 IEEE Globecom Workshops,
      Macau, Dec.~2026; the final version will appear in IEEE Xplore.}}}%
  \AtPageLowerLeft{\put(\LenToUnit{\dimexpr(\paperwidth-\textwidth)/2\relax},\LenToUnit{0.28in}){%
    \parbox[b]{\textwidth}{\scriptsize\copyright\,2026 IEEE. Personal use of this material
      is permitted. Permission from IEEE must be obtained for all other uses, in any current or
      future media, including reprinting/republishing this material for advertising or
      promotional purposes, creating new collective works, for resale or redistribution to
      servers or lists, or reuse of any copyrighted component of this work in other works.}}}%
}
\fi

\title{Toward Agentic AI for the 6G Physical Layer:
From Algorithm Discovery to Verified Baseband Silicon}

\input{glossaries}
\glsdisablehyper

\author{
\IEEEauthorblockN{
Eric Mittag\IEEEauthorrefmark{1}\IEEEauthorrefmark{3} \orcidlink{0009-0006-1318-332X},
Lukas Brechtel\IEEEauthorrefmark{1}\IEEEauthorrefmark{3} \orcidlink{0009-0008-9867-6653},
Maximilian Stark\IEEEauthorrefmark{2} \orcidlink{0000-0002-1750-5895},
Hans D. Schotten\IEEEauthorrefmark{1}\IEEEauthorrefmark{5} \orcidlink{0000-0001-5005-3635}
}
\IEEEauthorblockA{\IEEEauthorrefmark{1}German Research Center for Artificial Intelligence (DFKI), Kaiserslautern, Germany}
\IEEEauthorblockA{\IEEEauthorrefmark{2}CTO System Innovation, NXP Semiconductors, Hamburg, Germany}
\IEEEauthorblockA{\IEEEauthorrefmark{5}RPTU University Kaiserslautern-Landau, Germany}
\thanks{\IEEEauthorrefmark{3}These authors contributed equally.}
}
\maketitle

\begin{abstract}
6G will be designed faster than it can be built: AI-driven search already
proposes physical-layer algorithms at machine speed, while carrying a
single one onto embedded baseband silicon remains specialist work. That
silicon imposes hard memory and numeric limits, but also exposes cost
structures a GPU lacks, and a design flow that never runs on the target
sees neither. We present an agentic workflow in which an LLM coding
agent implements algorithms on a commercial baseband SoC, optimizes
them down to agent-written vector assembly, and validates them on silicon.
The agent is guided by a knowledge base of measured platform
facts, checked at every step against evaluators it is instructed never to edit,
and metered per phase. We showcase the workflow on an AI-discovered \gls{otfs} detector
portfolio on the open 6G4All platform. It yields twelve kernels that run
cycle-exactly on silicon, their optimized versions at 18\,k--405\,k cycles
(30--660\,$\boldsymbol{\mu}$s) per frame. It also yields a new detector shaped by
the chip's measured costs that needs 13.9\% fewer cycles than the port of the
discovered detector while matching its error rate, with a measured edge our
evaluation cannot statistically resolve, a structure that a hardware-blind
search objective had no reason to find. What is not specific to \gls{otfs} is the workflow: an LLM agent
exploring an implementation space under real hardware limits, with the cost
of every choice measured rather than assumed.
\end{abstract}


\glsresetall

\section{Introduction}
\label{sec:intro}

Whatever 6G turns out to be (new waveforms, integrated sensing, AI-native
receivers), each of its features will in the end be a
signal-processing algorithm that has to run inside the cycle, memory and
power envelope of baseband silicon. How fast those features reach real
radios will therefore be limited less by algorithm ideas than by the path
from idea to silicon. Candidate algorithms
already arrive faster than they can be deployed: \gls{llm}-driven evolutionary
search explores physical-layer solution spaces wholesale, and the framework of A\"it Aoudia
\emph{et~al.}~\cite{aite2026} explored 1114 distinct \gls{otfs} equalizer
architectures and produced a detector that outperforms published
baselines, in simulation and on \glspl{gpu}. Carrying any one of those ideas onto an
embedded baseband processor remains specialist work. Such parts earn their
power efficiency by omitting what a general-purpose core provides: the
\gls{dsp} used here has no scalar floating-point unit at all, so scalar
float arithmetic falls back on software emulation and costs orders of
magnitude more than the same work on the vector array.
Respecting that asymmetry is not an optimization detail; it decides whether
an algorithm fits at all. Agentic \gls{ai} has reached the physical layer at the level of intent and
configuration, in simulation~\cite{li2026agenticphy}; this paper turns it
toward implementation on the silicon itself: we
demonstrate, on the open 6G4All platform~\cite{mittag2026}, that an
appropriately instrumented environment
lets an \gls{llm} coding agent do that specialist work end to
end (implement, optimize at assembly level, validate on silicon) at
measured cost. If new algorithms can reach real baseband
hardware this cheaply, the hardware stops being where 6G research ends and
becomes part of the loop that proposes it.

A hardware-blind flow misses the platform in both directions. At its
published operating point the discovered detector's serialized channel
representation occupies some 63\,MB against roughly 28\,kB of usable target memory
(Table~\ref{tab:dfm}); no word length closes three orders of magnitude,
which is why this campaign runs at reduced grid dimensions, a consequence
of the platform rather than a convenience, and a constraint the algorithm
must be designed around rather than engineered away. And the platform's
\emph{advantages} are equally invisible to a search whose evaluator prices
only \gls{gpu} latency: that objective~\cite{aite2026} leaves memory and
\gls{dsp} arithmetic unpriced, so nothing rewards the structures this chip's cost
model does, which is where the campaign's new detector comes
from (Sec.~\ref{sec:novel}).

We report one campaign, run end to end with the \gls{otfs} portfolio as its
demonstration workload. It makes three contributions.
(i)~An agentic workbench in which every claim the agent makes is
falsifiable: measured platform knowledge, machine-checkable tooling, frozen
evaluators and a per-phase effort meter (Sec.~\ref{sec:workspace}).
(ii)~An executed showcase: four classical baselines and a discovered
detector re-derived clean-room, optimized in agent-written vector assembly
and validated on silicon, plus a new detector designed under the platform's
measured cost model that lands on the campaign's error-rate/cycle Pareto
front (Secs.~\ref{sec:campaign}--\ref{sec:results}).
(iii)~A per-phase accounting of engineering effort, metered from the agent's
own transcripts rather than self-reported (Sec.~\ref{sec:effort}).
A pre-registered, exploratory ablation probes the knowledge layer itself on two kernels
(Sec.~\ref{sec:ablation}): there, documentation changed not what the agent could
reach but how deep it went.

\section{Background and Related Work}
\label{sec:rlwrk}

\subsection{Platform and Source Framework}

The campaign runs on 6G4All~\cite{mittag2026}: a LimeSDR-Micro board around
the NXP LA9310, a commercial baseband \gls{soc} pairing a Cortex-M4 control
core with a second-generation \gls{vspa} vector \gls{dsp} (614.4\,MHz, 16
arithmetic units, up to 16 complex single-precision operations per clock).
The \gls{vspa} clock is exactly four times the part's maximum converter rate
of 153.6\,MS/s, so streaming work has a budget of four cycles per sample at
the maximum rate.
The working memories are hard limits: 16\,kB of \gls{dmem} (12\,160\,B usable for
static data), 32\,kB \gls{pram}, and 16\,kB of auxiliary \gls{ippu} data
memory used
as spill space. Firmware, drivers, kernel library and test infrastructure
are open repositories.

The discovery framework of~\cite{aite2026}, with its companion
study~\cite{aitaoudia2026telcoengineer}, couples an \gls{llm} that writes
candidate algorithms with an evolutionary archive and a Sionna-based
link-level evaluator~\cite{hoydis2022}. Its \gls{otfs} repository ships one
discovered artifact, an \gls{ec} detector with Doppler
block-diagonalization, Cholesky solves and a Richardson half-pass, plus
four tuned classical baselines: \gls{lmmse}, \gls{ep}~\cite{li2021ep},
Gaussian \gls{mp}~\cite{raviteja2018}, and \gls{uamp}~\cite{yuan2022uamp};
notably none is a neural network: the search produced structure plus
scalar constants, which is what an embedded target can afford. Its fitness is
computed entirely in simulation, modeling neither the cost of
transcendentals on a \gls{dsp} without scalar floating-point arithmetic nor
the
candidate's memory footprint (Table~\ref{tab:dfm}): no variant on the published
Pareto front is buildable as published.

\begin{table}[t]
\caption{Design rule check of the published channel representation against
the target: footprint of the sparse delay-Doppler view at 16-bit I/Q plus
16-bit index; \texttt{top\_k} is the number of taps kept per output row.
Row~1 is the framework's selected solution (NVE 0.549;
as shipped, \texttt{complex64}/\texttt{int64}, 2.7$\times$ more); row~2 is the
cheapest point on its published front, so the gap is not an artifact of
picking the largest variant. The last row is the simulator's linker map,
which exposes 8\,704\,B of the 16\,kB \gls{ippu}.}
\label{tab:dfm}
\centering
\footnotesize
\setlength{\tabcolsep}{4pt}
\begin{tabular}{lrrr}
\toprule
Grid & \texttt{top\_k} & Coefficients & Footprint \\
\midrule
$64 \times 64$ (selected soln.) & 2560 & $1.05 \cdot 10^{7}$ & 62.9\,MB \\
$64 \times 64$ (alt.\ Pareto pt.) & 1536 & $6.29 \cdot 10^{6}$ & 37.7\,MB \\
$8 \times 4$ (top-8 view)   &    8 & $2.56 \cdot 10^{2}$ & 1.5\,kB \\
\midrule
\multicolumn{3}{l}{Available on target (DMEM usable $+$ IPPU)} & 27.9\,kB \\
\multicolumn{3}{l}{\quad of which portable to the cycle simulator} & 20.4\,kB \\
\bottomrule
\end{tabular}
\end{table}

\subsection{Related Work}

End-to-end learned transceivers and learned pilotless
receivers~\cite{oshea2017, doerner2018, aitaoudia2022} optimize fixed
differentiable architectures at footprints far beyond embedded baseband
budgets; LLM-guided program search, covering
FunSearch~\cite{romeraparedes2024funsearch},
AlphaEvolve~\cite{novikov2025alphaevolve} and~\cite{aite2026} for the
\gls{phy}, closes its loop against software evaluators.

Classical word-length optimization~\cite{sung1995, constantinides2003, menard2006}
assigns precisions to a \emph{fixed} dataflow. Neither lever applies here:
this part computes in single precision only, so there are no word lengths to
choose, and closing the gap in Table~\ref{tab:dfm} needs the channel
recomputed rather than stored, a change of algorithm, not of precision.

Agentic implementation is an active line whose neighboring axes are
occupied: agents write low-level code for
general-purpose CPUs~\cite{mikek2026}, close loops against \emph{measured} hardware on
microcontrollers~\cite{zhang2026embeddedarena} and on \glspl{gpu} with
profiler feedback, reported cost and kernels no human wrote~\cite{zhang2025cudaforge,
luo2026agentickernel}, and carry
wireless functions to programmable radios, taking MATLAB through
\gls{hls} to bitstream and board with resource and latency
results~\cite{lei2026a2hmas}, or to verified over-the-air
emission~\cite{lei2026radiomaster}. \tii{Table~\ref{tab:related} places each of these on the axis it
occupies.}

Unoccupied is their conjunction on a fixed-function baseband part: assembly-level optimization against a vector
\gls{dsp}'s own \gls{isa}, gated by an external reference chain re-checked
on silicon, with per-phase effort reported as a result. The workbench described next is
built to close that gap.

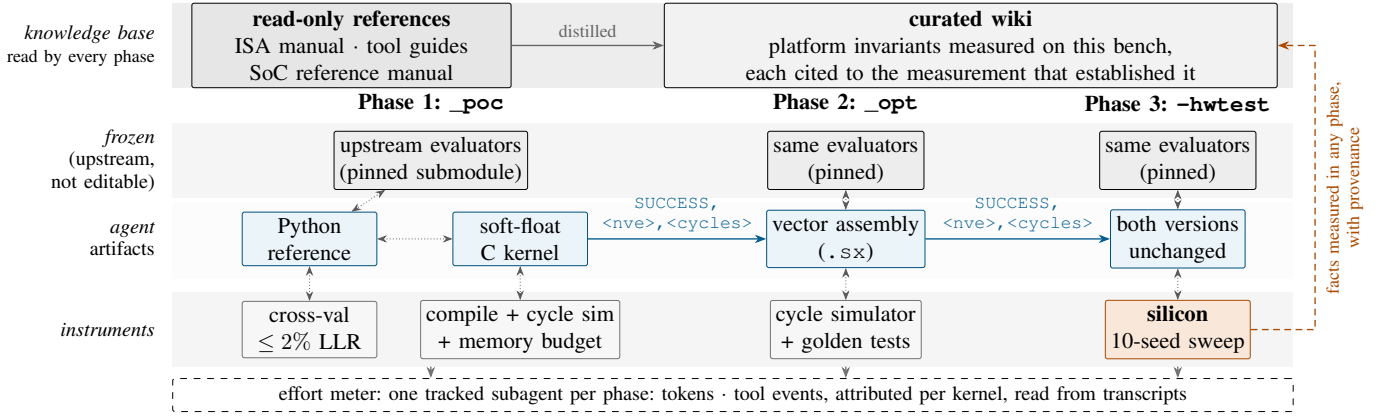
\begin{figure*}[t]
\centering
\resizebox{\textwidth}{!}{%

\begin{tikzpicture}[
  font=\normalsize,
  blk/.style={draw, rounded corners=1pt, align=center, inner sep=2.5pt,
              minimum height=1.55em, minimum width=6.0em, font=\normalsize},
  frozen/.style={blk, fill=black!7},
  agent/.style={blk, fill=optblue!8, draw=optblue!70!black},
  inst/.style={blk, fill=black!3, draw=black!55},
  si/.style={blk, fill=newverm!15, draw=newverm!80!black},
  lane/.style={font=\small\itshape, anchor=east, align=right},
  hdr/.style={font=\normalsize\bfseries, anchor=south},
  flow/.style={-{Stealth[length=1.8mm]}, semithick},
  chk/.style={{Stealth[length=1.5mm]}-{Stealth[length=1.5mm]}, densely dotted,
              black!65},
]

\def\cA{2.2} \def\cB{8.7} \def\cC{13.9}
\def\lK{4.75} \def\lF{2.95} \def\lA{1.72} \def\lI{0.30}
\def\R{17.6}

\begin{scope}[on background layer]
  \fill[black!7]  (0.05,\lK-0.66) rectangle (\R,\lK+0.66);
  \fill[black!4]  (0.05,\lF-0.58) rectangle (\R,\lF+0.58);
  \fill[black!1]  (0.05,\lA-0.62) rectangle (\R,\lA+0.58);
  \fill[black!4]  (0.05,\lI-0.58) rectangle (\R,\lI+0.58);
\end{scope}

\node[hdr] at (\cA+1.9,\lF+0.64) {Phase 1: \texttt{\_poc}};
\node[hdr] at (\cB+1.9,\lF+0.64) {Phase 2: \texttt{\_opt}};
\node[hdr] at (\cC+1.9,\lF+0.64) {Phase 3: \texttt{-hwtest}};

\node[lane] at (-0.1,\lK) {knowledge base\\\textnormal{\footnotesize read by every phase}};
\node[draw, rounded corners=1pt, fill=black!10, minimum width=5.0cm,
      minimum height=1.4em, anchor=west, align=center, font=\normalsize]
      (kbref) at (0.35,\lK)
  {\textbf{read-only references}\\ISA manual $\cdot$ tool guides\\SoC reference manual};
\node[draw, rounded corners=1pt, fill=black!5, minimum width=9.6cm,
      minimum height=1.4em, anchor=west, align=center, font=\normalsize]
      (kb) at (7.75,\lK)
  {\textbf{curated wiki}\\platform invariants measured on this bench,\\each cited to the measurement that established it};
\draw[flow, black!60] (kbref.east) -- node[above, font=\footnotesize, pos=0.5] {distilled} (kb.west);

\node[lane] at (-0.1,\lF) {frozen\\\textnormal{(upstream,}\\\textnormal{not editable)}};
\node[lane] at (-0.1,\lA) {agent\\\textnormal{artifacts}};
\node[lane] at (-0.1,\lI) {instruments};

\node[frozen] (ev1) at (\cA+1.9,\lF) {upstream evaluators\\(pinned submodule)};
\node[frozen] (ev2) at (\cB+1.9,\lF) {same evaluators\\(pinned)};
\node[frozen] (ev3) at (\cC+1.9,\lF) {same evaluators\\(pinned)};

\node[agent] (py)  at (\cA+0.0,\lA) {Python\\reference};
\node[agent] (cc)  at (\cA+3.3,\lA) {soft-float\\C kernel};
\node[agent] (sx)  at (\cB+1.9,\lA) {vector assembly\\(\texttt{.sx})};
\node[agent] (both)at (\cC+1.9,\lA) {both versions\\unchanged};

\node[inst] (xval) at (\cA+0.0,\lI) {cross-val\\$\le 2\%$ LLR};
\node[inst] (sim1) at (\cA+3.3,\lI) {compile + cycle sim\\+ memory budget};
\node[inst] (sim2) at (\cB+1.9,\lI) {cycle simulator\\+ golden tests};
\node[si]   (si)   at (\cC+1.9,\lI) {\textbf{silicon}\\10-seed sweep};

\draw[chk] (ev1) -- (py);
\draw[chk] (py)  -- (xval);
\draw[chk] (cc)  -- (sim1);
\draw[chk] (ev2) -- (sx);
\draw[chk] (sx)  -- (sim2);
\draw[chk] (ev3) -- (both);
\draw[chk] (both)-- (si);
\draw[chk] (py.east) -- (cc.west);

\draw[flow, optblue!80!black] ($(cc.east)+(0,0)$) -- node[above, font=\footnotesize,
  pos=0.5, align=center] {\texttt{SUCCESS,}\\\texttt{<nve>,<cycles>}} (sx.west);
\draw[flow, optblue!80!black] (sx.east) -- node[above, font=\footnotesize,
  pos=0.5, align=center] {\texttt{SUCCESS,}\\\texttt{<nve>,<cycles>}} (both.west);

\def\gutter{17.95}
\draw[flow, newverm!80!black, densely dashed]
  (si.east) -- (\gutter,\lI) -- (\gutter,\lK) -- (kb.east);
\node[font=\footnotesize, rotate=90, anchor=north, align=center, newverm!80!black]
  at (\gutter+0.06,{(\lI+\lK)/2}) {facts measured in any phase,\\with provenance};

\node[draw, dashed, rounded corners=1pt, minimum width=\R cm - 0.05cm,
      minimum height=1.2em, anchor=north west, font=\small,
      align=center] (meter) at (0.05,\lI-0.76)
  {effort meter: one tracked subagent per phase: tokens $\cdot$ tool
   events, attributed per kernel, read from transcripts};
\foreach \x in {4.1, 10.6, 15.8}
  \draw[flow, black!55] (\x,\lI-0.58) -- (\x,\lI-0.76);

\end{tikzpicture}%
}

\caption{The campaign as three metered phases. The agent writes its artifacts only in
the agent lane: the evaluators above are pinned upstream code it is instructed
not to edit, and the
instruments below bind every artifact to an executable reference. Silicon
appears once, at the end; facts measured there return to the knowledge base.}
\label{fig:loop}
\end{figure*}

\section{The Agentic Workbench}
\label{sec:workspace}

The campaign is run by a general-purpose \gls{llm} coding agent operating a
Linux bench host with no PHY-specific fine-tuning.\footnote{Claude Fable 5
at high reasoning effort for every metered phase; all reported figures come
from a single-model record.} What makes hardware
engineering work is not the model but the workbench around it: four
components, and a pipeline that decides when each instrument is used
(Fig.~\ref{fig:loop}).

The \emph{knowledge base} holds read-only references
(\gls{vspa} instruction-set manual, \gls{soc} reference manual, tool
guides) and a curated wiki of invariants measured on this bench, each cited
to its measurement. Facts the agent settles are written back with
provenance: 61 in this campaign, 43 of them reused by later phases, two
contradicting the \gls{isa} manual.

The \emph{instruments} make every step machine-checkable: the
vendor toolchain compiles C and \gls{vspa} assembly against a fixed linker
map whose overflow errors are themselves fitness signals; a cycle-accurate
simulator returns pass/fail and a cycle count for any kernel; and a
scripted debug bridge loads the kernel onto the core and reads the result
back, a source-to-silicon verdict in seconds without a boot flow or probe.

The \emph{campaign template} freezes what the agent must not touch: the discovery framework is
pinned as a git submodule and the harness imports its channel model,
\gls{ldpc} codec, \gls{qam} mapper and reference detectors call-for-call
(Fig.~\ref{fig:evalchain});
every candidate follows a fixed contract (faithful Python reference model,
float32 C or assembly kernel, golden vectors derived from the model,
self-checking tests) and is scored by one command whose first output line
is machine-parseable, \texttt{SUCCESS, <nve>, <cycles>}.

The \emph{effort meter} runs each phase as its own tracked subagent and reads tokens and
tool events from transcripts; nothing is self-reported.

\emph{Pipeline.} Every algorithm passes three gated phases
(Fig.~\ref{fig:loop}): \texttt{\_poc} produces a Python reference
(cross-validated where an upstream one exists) and a correctness-first C kernel, \texttt{\_opt} the
ISA-optimized version, and \texttt{-hwtest}, the only phase that touches
silicon, the validation. The ordering uses each instrument where it is
strong. Cross-validation against the upstream implementation and golden
vectors anchor semantics before any optimization exists. The headless simulator then carries the optimization
loop, where most iterations happen, and scales with compute rather than
benches: many candidates can be optimized in parallel against one board. Silicon appears once, at
the end, because the simulator proves to be an exact cost oracle
(Sec.~\ref{sec:silicon} finds the difference to be zero on all twelve
kernels), so hardware
validation confirms rather than discovers. Clean-room rules complete the protocol: the agent's only sources are the
template, the pinned checkout and the platform documentation, so measured
effort reflects derivation, not copying.

\section{Experimental Campaign}
\label{sec:campaign}

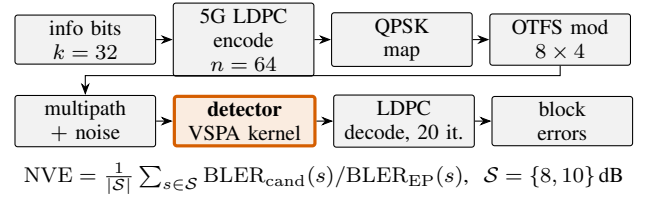
\begin{figure}[t]
\centering
\begin{tikzpicture}[node distance=0pt]
\tikzset{
  up/.style={blk, fill=black!5, minimum width=1.86cm, font=\footnotesize,
             minimum height=1.25em, inner sep=2pt},
  ag/.style={blk, fill=newverm!12, draw=newverm, very thick,
             minimum width=1.86cm, font=\footnotesize, minimum height=1.25em,
             inner sep=2pt},
  ar/.style={-{Stealth[length=1.6mm]}, thin},
}
\node[up] (bits)                          {info bits\\$k=32$};
\node[up, right=2.2mm of bits]  (enc)     {5G \gls{ldpc}\\encode\\$n=64$};
\node[up, right=2.2mm of enc]   (map)     {\gls{qpsk}\\map};
\node[up, right=2.2mm of map]   (mod)     {\gls{otfs} mod\\$8\times4$};
\foreach \a/\b in {bits/enc, enc/map, map/mod} \draw[ar] (\a) -- (\b);
\node[up, below=4.0mm of bits]  (chan)    {multipath\\$+$ noise};
\node[ag, right=2.2mm of chan]  (det)     {\textbf{detector}\\\gls{vspa} kernel};
\node[up, right=2.2mm of det]   (dec)     {\gls{ldpc}\\decode, 20 it.};
\node[up, right=2.2mm of dec]   (bler)    {block\\errors};
\foreach \a/\b in {chan/det, det/dec, dec/bler} \draw[ar] (\a) -- (\b);
\draw[ar] (mod.south) -- ++(0,-1.3mm) -| (chan.north);
\end{tikzpicture}

\vspace{0.2ex}
{\footnotesize$\mathrm{NVE}=\frac{1}{|\mathcal{S}|}\sum_{s\in\mathcal{S}}
  \mathrm{BLER}_{\mathrm{cand}}(s)/\mathrm{BLER}_{\mathrm{EP}}(s)$,
  \ $\mathcal{S}=\{8,10\}$\,dB}

\caption{The evaluation chain behind every \gls{nve}. Only the shaded block
is the agent's; every other stage is the framework's own
code~\cite{aite2026}, imported unmodified. Frames run in batches of eight
until 50 block errors accumulate, capped at 1600 (Sec.~\ref{sec:pareto}).}
\label{fig:evalchain}
\end{figure}

The workload comes from the discovery framework~\cite{aite2026} itself:
re-derive its four
classical \gls{otfs} baselines and its discovered \gls{ec} detector,
clean-room, as verified \gls{vspa} kernels, then design one new detector
under the platform's measured cost model. Dimensions follow from
Table~\ref{tab:dfm}: $M=8$ delay by $N=4$ Doppler bins, \gls{qpsk}, top-8
sparse channel view (top-20 for the \gls{ec} detectors, full channel for \gls{lmmse}), every
hyperparameter dimensionless. Before any C port
began, each baseline's reference model had to pass cross-validation against
the upstream torch implementation on identical frames (median relative
\gls{llr} deviation below 2\%, zero confident sign flips); all four passed
with median deviations of $1.3\cdot10^{-7}$ to $7.6\cdot10^{-7}$. The
discovered detector is gated by the coded \gls{nve} evaluation alone
(Fig.~\ref{fig:evalchain}), which resolves \gls{nve} differences only
coarsely (Sec.~\ref{sec:pareto}).

\subsection{ISA-Targeted Porting and Optimization}

With the workload and its gates fixed, the ports could begin. The
optimization gains are structural before they are instructional. The
block-circulant structure of the delay-Doppler channel along Doppler lets
$N$ independent $M\times M$ solves replace the dense $MN\times MN$ solve
(\texttt{lmmse\_opt}, 1693$\times$ over its proof of concept); for
\gls{qpsk} the \gls{mp} edge distributions factor exactly into I and Q,
collapsing each edge message from four log-values to two reals
(\texttt{mp\_opt}, 1066$\times$). In every optimized kernel the C driver
retains \emph{no} scalar floating-point arithmetic at all: transcendentals
are
polynomial chains on the vector array, and divisions use the
special-function unit's reciprocal. Where optimization changes numerics, the
contract requires the optimized version to carry its \emph{own} reference
model and re-earn its \gls{nve} rather than inherit it. All six did: five
exactly equal to their proof of concept, \texttt{ec2rich} within 0.006,
inside the protocol's resolution. The poc-to-opt ratios of 154--1693$\times$
(Table~\ref{tab:portfolio}) are therefore the priced distance between a
working kernel and a fast one, not the speedup of one artifact.

\subsection{Platform-Derived Detector Design: \texttt{ec\_frozen}}
\label{sec:novel}

The human brief asked only for one new detector of the agent's own,
shaped by what it had learned about the platform, and gated it on the
Pareto front; the frozen inverse, its stabilizers and the iteration count
below are the agent's choices, documented with their measured motivation in
the kernel's record. That record also lists the variants the agent measured
and rejected on its design-time two-point protocol: plain frozen Richardson
(\gls{nve} 1.83), the frozen step without Jacobi smoothing (3.17) or with
only the leading half (0.764), and four iterations instead of three (0.572
against 0.530). The campaign's own profiles supplied that motivation: in the
kernels that solve per-block systems the solves dominate, with eight Gauss-Jordan solves
at 38\% of \texttt{ec2rich\_opt}'s cycles, while the nonlinear projection
costs a fraction of a solve round. \texttt{ec\_frozen} exploits that imbalance. Like
the discovered detector it splits the channel into $N$ independent blocks
along Doppler, but it pays for each block's regularized inverse
\emph{once}, up front. Every subsequent \gls{ec} iteration reuses that
stale, frozen inverse only as a \emph{preconditioner}: instead of
refactorizing the block system, whose diagonal changes every iteration as
the site precisions evolve, it applies the frozen inverse and corrects the
result with a single Richardson step, a handful of matrix-vector products.
Naively this diverges once precisions grow (re-measured spectral radius up to 43);
three measured stabilizers make it work: the system is symmetrically
rescaled so the correction stays exact at both extremes of precision, each
frozen apply is bracketed by Jacobi smoothing steps, and every correction
starts from the previous iterate rather than from zero. Because a refinement costs 2.8\,k cycles
against 7.6\,k for a factorized solve round, a third \gls{ec} iteration
becomes affordable. It closes with a linear-collapsed \gls{qpsk} demapper and
is gated by reaching the Pareto front of the campaign's own ports.
Algorithm~\ref{alg:ecfrozen} gives the procedure and its main
hyperparameters.

\begin{algorithm}[t]
\caption{\texttt{ec\_frozen} per frame. $\mathbf{H}_n$, $\mathbf{y}_n$:
block $n$ of the channel and of the received grid after $\mathbf{F}$, the
unitary length-$N$ \gls{dft} across Doppler; $\hat\sigma^2$: noise variance
inflated by 1.15; $\mathbf{z}_n$: estimate in block $n$. \gls{ec} site:
precision $\boldsymbol{\pi}$, precision-weighted mean $\boldsymbol{\eta}$. $\mathbf{x}_{\mathrm{lin}}$,
$\mathbf{v}$: linear estimate, its variance; $\mathbf{p}$: posterior precision;
$\mathbf{p}_A$, $\mathbf{r}_A$: extrinsic precision, mean; $\hat{\mathbf{x}}$,
$\mathbf{p}_{\mathrm{nl}}$: \gls{qpsk} posterior mean, its precision;
$\mathbf{s}$, $\mathbf{w}$: rescale, Jacobi weights. Vectors are indexed by
delay bin ($\boldsymbol{\eta}$, $\hat{\mathbf{x}}$, $\mathbf{x}_{\mathrm{lin}}$,
$\mathbf{r}_A$ span the grid); products, quotients and $(\cdot)^{-1/2}$ act
elementwise; $\mathrm{diag}$ forms or extracts a diagonal; $[\cdot]$ clips to
$[p_{\max}^{-1},p_{\max}]$. $T{=}3$ \gls{ec} iterations, Jacobi weight
$\omega{=}0.8$, damping $\delta{=}0.68$, precision clamp $p_{\max}{=}64$.}
\label{alg:ecfrozen}
\footnotesize
\begin{algorithmic}[1]
\Require per block $n$: $\mathbf{G}_n=\mathbf{H}_n^{\mathsf{H}}\mathbf{H}_n/\hat\sigma^2$,
  $\mathbf{b}_n=\mathbf{H}_n^{\mathsf{H}}\mathbf{y}_n/\hat\sigma^2$
\State $\mathbf{B}_n \gets (\mathbf{G}_n+\mathbf{I})^{-1}$ for all $n$
  \Comment{the only inverses}
\State $\boldsymbol{\pi}\gets\mathbf{1}$, $\boldsymbol{\eta}\gets\mathbf{0}$,
  $\mathbf{z}_n\gets\mathbf{B}_n\mathbf{b}_n$ \Comment{exact at $\boldsymbol{\pi}=\mathbf{1}$}
\For{$t=1,\dots,T$}
  \If{$t>1$} $\mathbf{z}\gets\Call{FrozenSolve}{\boldsymbol{\pi},\boldsymbol{\eta},\hat{\mathbf{x}}}$ \EndIf
  \State $\mathbf{x}_{\mathrm{lin}}\gets\mathbf{F}^{\mathsf{H}}\mathbf{z}$;
    $\mathbf{v}\gets$ mean over $n$ of $\mathrm{diag}\,\mathbf{B}_n/(\mathbf{1}+(\boldsymbol{\pi}-\mathbf{1})\,\mathrm{diag}\,\mathbf{B}_n)$
  \State $\mathbf{p}\gets\min(1/\mathbf{v},p_{\max})$, $\mathbf{p}_A\gets[\mathbf{p}-\boldsymbol{\pi}]$,
    $\mathbf{r}_A\gets(\mathbf{p}\,\mathbf{x}_{\mathrm{lin}}-\boldsymbol{\eta})/\mathbf{p}_A$
  \State $\hat{\mathbf{x}}\gets \kappa\tanh(2\kappa\,\mathbf{p}_A\mathbf{r}_A)$ per component,
    $\kappa=2^{-1/2}$; $\mathbf{p}_{\mathrm{nl}}\gets$ its precision, averaged over Doppler
  \State $\boldsymbol{\eta}\gets(1-\delta)\boldsymbol{\eta}+\delta(\mathbf{p}_{\mathrm{nl}}\hat{\mathbf{x}}-\mathbf{p}_A\mathbf{r}_A)$
  \State $\boldsymbol{\pi}\gets[(1-\delta)\boldsymbol{\pi}+\delta\,[\mathbf{p}_{\mathrm{nl}}-\mathbf{p}_A]]$
\EndFor
\State one more \textsc{FrozenSolve}, then the collapsed \gls{qpsk} demap
\Function{FrozenSolve}{$\boldsymbol{\pi},\boldsymbol{\eta},\hat{\mathbf{x}}$}
  \Comment{each block $n$}
  \State $\mathbf{A}_n\gets\mathbf{G}_n+\mathrm{diag}\,\boldsymbol{\pi}$,
    $\mathbf{c}_n\gets\mathbf{b}_n+\boldsymbol{\pi}\,(\mathbf{F}(\boldsymbol{\eta}/\boldsymbol{\pi}))_n$
  \State $\mathbf{s}\gets(\mathbf{1}+(\boldsymbol{\pi}-\mathbf{1})\,\mathrm{diag}\,\mathbf{B}_n)^{-1/2}$,
    $\mathbf{w}\gets\omega/(\mathrm{diag}\,\mathbf{G}_n+\boldsymbol{\pi})$
  \State $\mathbf{z}_n\gets(\mathbf{F}\hat{\mathbf{x}})_n$ \Comment{warm start}
  \State $\mathbf{z}_n\gets\mathbf{z}_n+\mathbf{w}(\mathbf{c}_n-\mathbf{A}_n\mathbf{z}_n)$ \Comment{Jacobi}
  \State $\mathbf{z}_n\gets\mathbf{z}_n+\mathbf{s}\,\mathbf{B}_n\big(\mathbf{s}(\mathbf{c}_n-\mathbf{A}_n\mathbf{z}_n)\big)$
    \Comment{frozen}
  \State $\mathbf{z}_n\gets\mathbf{z}_n+\mathbf{w}(\mathbf{c}_n-\mathbf{A}_n\mathbf{z}_n)$ \Comment{Jacobi}
\EndFunction
\end{algorithmic}
\end{algorithm}

\begin{table}[t]
\caption{The OTFS kernel portfolio. Cycles per frame, identical in
simulator and on silicon (Sec.~\ref{sec:silicon}); NVE at the protocol's
resolution, with approximate 95\% intervals of about $\pm$0.15
(Sec.~\ref{sec:pareto}). $\star$ designed in this campaign.}
\label{tab:portfolio}
\centering
\footnotesize
\setlength{\tabcolsep}{2.6pt}
\begin{tabular}{lrrrrr}
\toprule
Kernel & NVE & Cycles & vs.\ PoC & DMEM & PRAM \\
\midrule
\texttt{lmmse\_poc}  & 0.65 & 31\,008\,649 & --- & 69.5\% & 49.0\% \\
\texttt{lmmse\_opt}  & 0.65 & \textbf{18\,318} & 1693$\times$ & 97.9\% & 50.5\% \\
\texttt{ep\_poc}     & 0.94 & 62\,503\,987 & --- & 76.9\% & 51.0\% \\
\texttt{ep\_opt}     & 0.94 & 405\,379 & 154$\times$ & 100\% & 70.5\% \\
\texttt{mp\_poc}     & 1.09 & 248\,238\,967 & --- & 83.6\% & 53.1\% \\
\texttt{mp\_opt}     & 1.09 & 232\,847 & 1066$\times$ & 95.8\% & 54.2\% \\
\texttt{uamp\_poc}   & 0.94 & 48\,598\,151 & --- & 82.4\% & 83.5\% \\
\texttt{uamp\_opt}   & 0.94 & 266\,422 & 182$\times$ & 98.9\% & 95.8\% \\
\texttt{ec2rich\_poc}& 0.57 & 6\,389\,937 & --- & 90.0\% & 87.2\% \\
\texttt{ec2rich\_opt}& 0.56 & 39\,520 & 162$\times$ & 98.9\% & 88.8\% \\
\texttt{ec\_frozen\_poc}$^\star$ & 0.51 & 9\,710\,079 & --- & 97.3\% & 81.1\% \\
\texttt{ec\_frozen\_opt}$^\star$ & \textbf{0.51} & \textbf{34\,010} & 286$\times$ & 98.9\% & 88.0\% \\
\bottomrule
\end{tabular}
\end{table}

\section{Results}
\label{sec:results}

\subsection{Silicon Validation}
\label{sec:silicon}

All twelve kernels were promoted to silicon in the separated hardware
phase, over a self-hosted transport that needs no debug probe.
\textbf{Every kernel reproduced its simulator cycle count exactly}, with
zero difference, across four orders of magnitude of frame cost, soft-float C and
agent-written assembly, 0--8704\,B of \gls{ippu} data memory (in bytes, since
simulator and chip expose different ceilings), and up to 100\%
\gls{dmem} occupancy; at 248\,M cycles that is agreement to one part in
$2.5\cdot10^{8}$. A smoke kernel shows a reproducible one-cycle offset, so the instrument
resolves a single cycle and the twelve exact matches are not measurement
insensitivity. The agreement is a
property of the code, not of one trace: most detectors have data-dependent
cycle counts (\texttt{mp\_poc} varies 2.8$\times$ across seeds), ten-seed
sweeps, each seed a fresh channel, noise and payload draw, exercise those
paths on every kernel. A seed passes when simulator and silicon
cycle counts are equal, the \glspl{llr} meet the golden tolerance
(absolute $5\cdot10^{-3}$, relative $2\cdot10^{-3}$), and every \gls{llr}
the reference is confident about ($|\mathrm{LLR}|>1$) keeps its sign:
agreement is exact in cycles, toleranced in values and exact in confident
bit decisions. Across 120
seed runs: 113 pass, 6 are generator margin-guard rejections
(\texttt{ep\_opt} seeds 101 and 108, both \gls{mp} kernels' seeds 102 and
109; re-running their host generators reproduces each rejection, so none is
a masked dependency failure), and one fails, the \texttt{uamp\_opt} seed discussed below. Two mechanisms by which
silicon might still have diverged were falsified. The same workspace is
placed roughly 500\,kB apart in the simulator and chip address maps yet
costs identical cycles, so placement does not affect timing; and kernels at
100\% \gls{dmem} occupancy match as exactly as those at 69.5\%, so the chip
adds no memory wait states the simulator omits.

\label{sec:guardrails}
What the hardware phase caught was faults in the \emph{verification
tooling}, not the silicon, in two ways. First, the multi-seed sweep script silently classified missing
Python dependencies as margin-guard skips and returned success; a table of
skipped seeds was indistinguishable from a clean pass. Second, the frame
generators' float32 twin cannot see the assembly kernels' special-function
arithmetic: one \texttt{uamp\_opt} seed misses the golden tolerance by
1.4$\times$, bit-identically on silicon and in the simulator and with no
confident sign flipped, while the twin that screens seeds scores that frame
at 0.1\% of its budget. We keep and report the seed. The fixes are a
sweep that tells a failed generator from a guard rejection, and a twin that
emulates the special-function approximations, as several kernels' reference
models already do for their exponential, instead of exact float32. This is the sim-to-real lesson displaced one level: on a
silicon-exact simulator the residual reality gap moves out of the hardware
and into the verification models.

\subsection{The Pareto Front and the Frame Budget}
\label{sec:pareto}

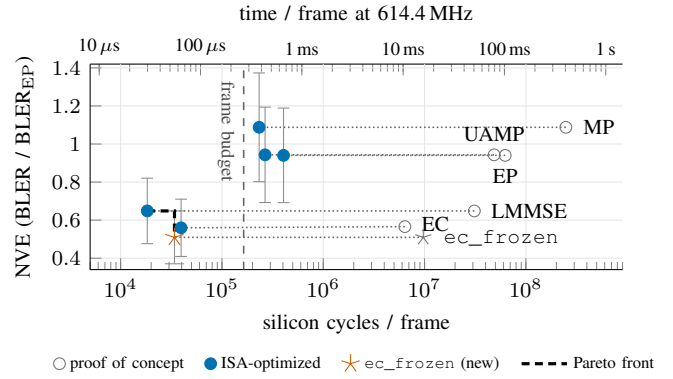
\begin{figure}[t]
\centering
\begin{tikzpicture}
\begin{axis}[
  name=timeaxis,
  width=0.98\columnwidth, height=4.3cm,
  xmode=log, xmin=8.138, xmax=1.4648e6,
  ymin=0, ymax=1, ytick=\empty,
  axis x line*=top, axis y line=none,
  xlabel={time / frame at 614.4\,MHz},
  xtick={10,100,1000,10000,100000,1000000},
  xticklabels={10\,$\mu$s,100\,$\mu$s,1\,ms,10\,ms,100\,ms,1\,s},
  tick label style={font=\scriptsize},
  label style={font=\footnotesize},
]
\end{axis}
\begin{axis}[
  at={(timeaxis.south west)}, anchor=south west,
  width=0.98\columnwidth, height=4.3cm,
  xlabel={silicon cycles / frame}, ylabel={NVE (BLER / BLER$_{\mathrm{EP}}$)},
  xmode=log, xmin=5e3, xmax=9e8, ymin=0.34, ymax=1.42,
  ytick={0.4,0.6,0.8,1.0,1.2,1.4},
  ylabel near ticks, xlabel near ticks,
  axis x line*=bottom, axis y line*=left,
  tick label style={font=\footnotesize},
  label style={font=\footnotesize},
  legend style={font=\scriptsize, at={(0.5,-0.37)}, anchor=north,
                draw=none, legend columns=4,
                /tikz/every even column/.append style={column sep=6pt},
                nodes={scale=0.9}, row sep=-1pt},
  grid=major, grid style={black!10},
]

\addplot[dashed, semithick, black!60, forget plot] coordinates {(163840,0.34)(163840,1.42)};
\node[font=\scriptsize, anchor=north west, rotate=-90, black!70]
  at (axis cs:163840,1.38) {frame budget};

\addplot[densely dotted, semithick, black!60, forget plot] coordinates {(31008649,0.6483)(18318,0.6483)};
\addplot[densely dotted, semithick, black!60, forget plot] coordinates {(62503987,0.9403)(405379,0.9403)};
\addplot[densely dotted, semithick, black!60, forget plot] coordinates {(248238967,1.0879)(232847,1.0879)};
\addplot[densely dotted, semithick, black!60, forget plot] coordinates {(48598151,0.9431)(266422,0.9431)};
\addplot[densely dotted, semithick, black!60, forget plot] coordinates {(6389937,0.5658)(39520,0.5596)};
\addplot[densely dotted, semithick, black!60, forget plot] coordinates {(9710079,0.5095)(34010,0.5095)};

\addplot[only marks, mark=star, mark size=3.2pt, black!55, forget plot] coordinates {
  (9710079,0.5095)
};
\addplot[only marks, mark=o, mark size=2.2pt, black!55] coordinates {
  (31008649,0.6483)(62503987,0.9403)(248238967,1.0879)(48598151,0.9431)
  (6389937,0.5658)
};
\addlegendentry{proof of concept}
\addplot[only marks, mark=*, mark size=2.2pt, optblue,
  error bars/.cd, y dir=both, y explicit,
  error bar style={black!45, thin}] coordinates {
  (18318,0.6483) +- (0,0.1721) (405379,0.9403) +- (0,0.2486)
  (232847,1.0879) +- (0,0.2857) (266422,0.9431) +- (0,0.2507)
  (39520,0.5596) +- (0,0.1506)
};
\addlegendentry{ISA-optimized}
\addplot[only marks, mark=star, mark size=3.6pt, newverm,
  error bars/.cd, y dir=both, y explicit,
  error bar style={black!45, thin}] coordinates {
  (34010,0.5095) +- (0,0.1390)
};
\addlegendentry{\texttt{ec\_frozen} (new)}
\addplot[densely dashed, very thick, black] coordinates {(18318,0.6483)(34010,0.6483)(34010,0.5095)};
\addlegendentry{Pareto front}

\node[font=\footnotesize, anchor=west] at (axis cs:3.7e7,0.6483) {LMMSE};
\node[font=\footnotesize, anchor=south] at (axis cs:4.86e7,0.965) {UAMP};
\node[font=\footnotesize, anchor=north] at (axis cs:6.25e7,0.918) {EP};
\node[font=\footnotesize, anchor=west] at (axis cs:3.0e8,1.0879) {MP};
\node[font=\footnotesize, anchor=west] at (axis cs:7.6e6,0.580) {EC};
\node[font=\footnotesize, anchor=west] at (axis cs:1.25e7,0.5095) {\texttt{ec\_frozen}};
\end{axis}
\end{tikzpicture}
\caption{Coded error rate versus measured silicon cycles per frame; the top
axis converts cycles to time, the dashed vertical marks the 267\,$\mu$s frame
budget.
Dotted ties link each algorithm's two versions (only
\texttt{ec2rich}'s is non-horizontal, 0.006 apart).
The black dashed staircase is
the Pareto front of the measured values.
Vertical bars: approximate 95\% \gls{nve} intervals of the optimized kernels
(Sec.~\ref{sec:pareto}); cycles are exact. The three kernels near the front
overlap in \gls{nve}; \gls{ep}, \gls{uamp} and \gls{mp} lie above them and are
dominated by \texttt{ec\_frozen\_opt} even at the interval level.}
\label{fig:pareto}
\end{figure}

Fig.~\ref{fig:pareto} shows the result. The optimization phase moves each
algorithm two to three decades or more left at unchanged \gls{nve}; the surviving
front has two points, \texttt{lmmse\_opt} at 18\,318 cycles and the new
\texttt{ec\_frozen\_opt} at 34\,010, so 1.9$\times$ the cycles buys a measured 0.14 of
\gls{nve} (0.65 against 0.51), itself within the intervals of
Fig.~\ref{fig:pareto}. Against our port of the framework's own discovered detector,
\texttt{ec\_frozen\_opt} needs 34\,010 against 39\,520 cycles, a
deterministic 13.9\% reduction, at a measured \gls{nve} of 0.51 against 0.56.

That \gls{nve} margin, however, is not resolved by this protocol. The
protocol is the framework's own Monte-Carlo setting (batches of eight, 50
block errors or at most 1600 frames per point), kept unmodified and frozen before the campaign so
that no agent could tune the gate it is scored by; it separates detectors
that differ by tens of percent, as \gls{ep}, \gls{uamp} and \gls{mp} do from
\texttt{ec\_frozen\_opt} (Fig.~\ref{fig:pareto}), but not the front's members from each
other. The evaluation stores rates, not counts. The stopping rule lets the counts be
recovered uniquely, and the same procedure reproduces the reduction sweep's
recorded counts in all 50 cases. Treating the curtailed sampling as
negative-binomial then gives approximate 95\% intervals of $\pm$0.14 and
$\pm$0.15, about threefold the separation. Over all six stored \gls{snr}
points (2--12\,dB) neither detector is consistently ahead (equal counts at
2 and 4\,dB, \texttt{ec\_frozen\_opt} lower at 8 and 10\,dB,
\texttt{ec2rich\_opt} at 6 and 12\,dB), and every pair's 95\%
Clopper--Pearson intervals overlap. The runs shared frames only up to
8\,dB, since each stopping point shifts the random stream, so the records
allow no paired comparison. The defensible claim is therefore its place on the
front: strictly cheaper at an error rate this protocol cannot separate, that
is, \gls{nve} parity rather than superiority. Settling the error-rate
difference needs a paired re-evaluation that retains per-frame outcomes. Cycle counts carry no such caveat; they are
deterministic and exact.

The counts are also affordable: the frame spans
$N/\Delta f = 267\,\mu\mathrm{s}$ at the upstream subcarrier spacing
$\Delta f{=}15$\,kHz, a budget of
roughly 164\,k cycles at 614.4\,MHz, and both front points sit inside
it (\texttt{lmmse\_opt} at 11\%, \texttt{ec\_frozen\_opt} at 21\%, the
discovered detector's port at 24\%). Both front kernels are free of
data-dependent branches and returned the same count on every seed, so these
are worst-case figures, not averages. \texttt{mp\_opt}, \texttt{uamp\_opt} and
\texttt{ep\_opt} exceed it by 1.4$\times$, 1.6$\times$ and 2.5$\times$. The reduction
that forces the small grid also shortens the frame, so this is a statement
about the operating point; what it establishes is that the front's detection
kernel fits the frame budget on this target with margin. The counter runs
from the call, with inputs already in \gls{dmem}, to the return,
so moving a frame in and its LLRs out is not included.

\subsection{Engineering Effort}
\label{sec:effort}

Table~\ref{tab:effort} is the campaign's effort record: agent output tokens
per phase (including billed reasoning tokens), deduplicated per API
response. It counts the metered phases only, excluding the coordinating
sessions, and each phase optimizes once to a depth the agent chooses: these
are not the tokens to reach a best kernel. Tokens are the only effort
metric we report: monetary cost moves with prices and caching
discounts, and would not be reproducible from the archive. Tokens omit
wall-clock time and the human preparation of the workbench. Human input to
each phase was its task brief; in the seven phases whose transcripts are
retained, the only further input was eleven continuation prompts without
technical content. Three observations follow.

\emph{Deriving an algorithm is cheap and uniform; making it fast is not:}
across the five ports the proof-of-concept phases cost 219.5\,k output
tokens and the optimization phases 1604.5\,k, a 7.3$\times$ asymmetry, and
proof-of-concept cost is flat (38.5--52.5\,k) whether the algorithm is a
one-shot linear solve (\gls{lmmse}) or a fifteen-iteration message-passing
scheme (\gls{mp}).

\emph{Silicon is the cheapest phase:} validating all twelve kernels cost
97.0\,k tokens, 4\% of the metered total, a consequence of the simulator
being exact, inverting the assumption that hardware is the scarce tier.

\emph{Documented knowledge is reused across phases:} the final optimization phase,
re-run blind on a stripped repository, passed its golden test on the
\emph{first} simulator run and built no probe kernels, having inherited four candidates' documented
\gls{isa} facts (an audit traces 43 of 50 earlier facts into reuse).

The phases ran in sequence by design, so
per-algorithm effort is a point on a learning trajectory, not an
independent sample.

\begin{table}[t]
\caption{Measured engineering effort per algorithm and phase: agent output
tokens ($10^3$, incl.\ billed reasoning), deduplicated per API response.}
\label{tab:effort}
\centering
\footnotesize
\setlength{\tabcolsep}{4.5pt}
\begin{tabular}{lrrrr}
\toprule
 & \multicolumn{3}{c}{Output tokens ($10^3$)} & \\
\cmidrule(lr){2-4}
Algorithm & PoC & Opt & HW & Total \\
\midrule
LMMSE            & 38.5 & 228.5 & 12.5 & 279.5 \\
EP               & 41.2 & 387.4 & 12.5 & 441.0 \\
MP               & 39.7 & 249.4 & 14.7 & 303.7 \\
UAMP             & 47.7 & 432.0 & 32.4 & 512.0 \\
EC (discovered)  & 52.5 & 307.2 & 13.5 & 373.2 \\
\texttt{ec\_frozen} (new) & 83.7 & 241.9 & 11.4 & 337.0 \\
\midrule
18 metered phases & 303.2 & 1846.4 & 97.0 & 2246.6 \\
\bottomrule
\end{tabular}
\end{table}

\subsection{Ablation of the Knowledge Layer}
\label{sec:ablation}

Repeating the \gls{lmmse} port clean-room reproduced the campaign closely
(same \gls{nve}; 18\,111 against 18\,318 cycles), suggesting its operating
points are set by the protocol rather than by one run. That control makes a
stronger experiment possible: whole phases re-run with the knowledge layer
reduced or removed\tii{ (Table~\ref{tab:ablation})}, pre-registered with token
caps and outcome criteria fixed in advance and withheld from the workers,
which saw only their task brief, on \gls{lmmse}, the
portfolio's simplest port and the only one whose campaign run started from
the same knowledge state as the documented arm, and on \gls{ep}, the port that leans hardest on
hand-built approximations: it needs a vector exponential and a log-sum-exp
the \gls{isa} does not provide, and runs at 100\% \gls{dmem}. The two
bracket the portfolio by design; each arm re-runs whole phases at
0.24--0.46\,M output tokens, which is why the ablation covers these two
rather than all six and runs each cell once.

Removing documentation changed optimization tokens-to-SUCCESS by 0.71$\times$
(\gls{lmmse}) and 1.07$\times$ (\gls{ep}), and every \gls{lmmse}
arm found the block-diagonal mathematics unaided, the
documentation-free one first, in its proof of concept. What changed was
depth: the manuals-only arm reached a slightly faster kernel
than the wiki-plus-manuals arm (16\,728 against 18\,111 cycles), while the two
documentation-free arms stayed largely in C with the SDK's intrinsics
instead of descending to hand-scheduled assembly, and settled
2.2$\times$ and 1.8$\times$ slower. The brief invited assembly where it paid
and gated only on improving over the proof of concept, identically in every
arm. The effect is consistent with the
\gls{isa} documentation carrying it, since manuals alone sufficed, rather
than with the campaign's accumulated fact base. The ablation scores the
knowledge layer only by what the agent produced. It does not measure the
second function: facts written back with provenance are what let a human
audit a campaign at all, including the reuse audit above and the two entries
that contradict the \gls{isa} manual.

With one run per cell and kernels that are not accuracy-matched, since each
arm re-earned its \gls{nve} in its own environment, this is exploratory and
holds for these two kernels only. Gates and tooling
alone were enough to carry even the documentation-free arms to correct
kernels.

\begin{table}[t]
\caption{Link cost of the grid reduction: coded BLER-$10^{-1}$ crossings of
the upstream detectors through one chain at both operating points: lite
($8\times4$, \gls{qpsk}, codeword $n{=}64$) and full ($64\times64$, 16-\gls{qam},
$n{=}16\,384$; 3- vs 6-path channel). Pre-registered sweep, $\geq$50 block
errors or 1600 frames per point; $\Delta_{\mathrm{red}}$ = full $-$ lite.}
\label{tab:redcost}
\centering
\footnotesize
\setlength{\tabcolsep}{5pt}
\begin{tabular}{lrrr}
\toprule
Detector & lite (dB) & full (dB) & $\Delta_{\mathrm{red}}$ (dB) \\
\midrule
UAMP            & 8.55 & 12.54 & $+3.98$ \\
MP              & 9.03 & 13.14 & $+4.11$ \\
EC (discovered) & 7.92 & 12.30 & $+4.38$ \\
EP              & 8.58 & 13.08 & $+4.50$ \\
LMMSE           & 7.82 & 14.55 & $+6.73$ \\
\bottomrule
\end{tabular}
\end{table}

\section{Discussion and Limitations}
\label{sec:disc}

\emph{What does the forced reduction cost?} A pre-registered sweep of the
upstream detectors through one chain at both operating points measures
it (Table~\ref{tab:redcost}): the full protocol's BLER-$10^{-1}$
crossings sit 4.0--4.5\,dB above the lite protocol's for four
detectors but
6.7\,dB for \gls{lmmse}, quantifying the ranking inversion behind the
front. The claim is therefore not that a designed detector beats a classical
one at this size, but that the loop produced a front-defining design
\emph{in the regime that favors the simple linear solve most}, at a
measured cost.

\emph{Why the search missed it.} The cost model that makes
\texttt{ec\_frozen} attractive does not exist on a \gls{gpu}, where
factorization is not the bottleneck: the framework's own
Krylov/preconditioned family drew 130 candidates and scored second-worst of
eleven~\cite{aite2026}. We do not claim the search could not have found this detector; its
objective gave it no gradient toward it. The remedy is analytical and cheap,
a design rule check at search time (footprint against data memory, code
against program memory, operations against a slot budget) which this campaign
calibrates with the twelve kernels' measured silicon cycles.

\emph{Limitations.} The campaign's results hold for one operating point: $8\times4$
(Table~\ref{tab:dfm}), \gls{qpsk}, \gls{nve} at two \gls{snr} points; \gls{nve} is normalized to each protocol's own
\gls{ep} baseline, so our values are not comparable with the framework's
published ones, and dominance at $64\times64$ is not established. Scaling
is not demonstrated: the Doppler-block decomposition makes the $N$ blocks
independent, so a frame can stream block by block through double-buffered
\gls{dma}, but at $64\times64$ one block's Gram matrix alone (32\,kB in
single precision) exceeds \gls{dmem}, so tiling must also reach inside the
block, or the workload must move to a larger member of the family; no kernel has run in a live stack; and one
optimized kernel misses the golden tolerance on one seed in ten, identically
in simulator and silicon and with no bit decision changed
(Sec.~\ref{sec:silicon}). The ablation covers two of six
algorithms with one run per cell (Sec.~\ref{sec:ablation}), its arms still
saw the vendor SDK's example assembly, and ablating the full campaign remains open.
Silicon validation covers correctness, memory fit and cycles; power, energy
per frame and the \gls{dma} cost of moving a frame in and out were not
measured. No non-agentic flow (compiler with hand-written intrinsics,
autotuning, or an engineer under the same budget) was run as a baseline;
the nearest point is the ablation's documentation-free arms, which stayed
largely in C with the SDK's intrinsics, 1.8--2.2$\times$ slower than assembly
(Sec.~\ref{sec:ablation}). The campaign therefore establishes feasibility at
a measured agent cost, not a productivity gain over such flows.

\section{Conclusion and Outlook}
\label{sec:concl}

This work showed, for one detector portfolio on one commercial baseband
\gls{soc}, that the step from a discovered algorithm to optimized silicon
can be automated. Inside an instrumented workbench of
measured platform knowledge, machine-checkable tooling and evaluators frozen
by protocol, an \gls{llm} agent carried a discovered \gls{otfs} portfolio
to verified, \gls{isa}-optimized kernels, with no vector assembly written by hand.

The result holds where it can be checked. All twelve kernels reproduce their
simulator cycle counts exactly on silicon; the optimized versions run at
18\,k--405\,k cycles (30--660\,$\mu$s) per frame, and the two kernels on the
resulting error-rate/cycle front spend 11\% and 21\% of the 267\,$\mu$s
frame. A detector designed under the platform's measured cost model, reusing
one frozen inverse instead of refactorizing, needs 13.9\% fewer cycles than
the discovered detector's port while matching its error rate, with a measured
edge the protocol cannot statistically resolve, a structure the source
search objective did not reward. The campaign also measures how that effort was distributed: of 2.25\,M metered agent tokens, 14\% went to proof of concept, 82\% to
assembly optimization and 4\% to hardware.

Nothing in the workflow is specific to \gls{otfs}, although only \gls{otfs}
detection has been tested: what the campaign demonstrates is a
way of working, with measured platform cost inside the loop that writes the
code and every claim checked against something the agent is not permitted to
edit. That is the
direction 6G4All is built for: baseband hardware as a place
where new algorithms are routinely tried, not only finally shipped. Because
the platform is inexpensive commodity hardware, the environment scales
outward: next are an over-the-air testbed running agent-produced kernels
under real channels, and lab instrumentation (vector signal analysis, power,
controlled \gls{rf}) so the agent sees every \gls{rf}-relevant property, not
only cycles and bit-exactness. Larger members of the vendor's baseband line
(up to eight \gls{vspa} cores) admit the grids where iterative detectors
earn their cost back. A
discovery run under a hardware-aware objective is the experiment this work
sets up.

\ifackfunding\aidisclosuretrue\fi
\ifaidisclosure\section*{Acknowledgment}\fi
\ifackfunding
This work has been supported by the Federal Ministry of Research, Technology
and Space of the Federal Republic of Germany (grant agreements 16KIS2402K,
Open6GHub+, and 16KISS008, XCOM).

\fi
\ifaidisclosure
\emph{Use of AI-generated content.} Anthropic Claude Fable 5 generated the
\gls{vspa} artifacts studied in Sections~\ref{sec:workspace}--\ref{sec:results}
and, together with Claude Opus 5 and Opus 5.5, materially assisted the organization,
drafting and revision of all manuscript sections, figures and tables. The
authors critically reviewed and edited all AI-generated text, checked the
technical claims and reported quantities against source code, tests and
archived measurement records, approved the final manuscript, and take full
responsibility for its accuracy, originality and integrity.
\fi

\printbibliography

\ifdraftmarks\clearpage\fi
\openitems

\end{document}

%% file: glossaries.tex
\newacronym{5g}{5G}{fifth generation}
\newacronym{6g}{6G}{sixth generation}
\newacronym{newr}{NR}{New Radio}
\newacronym{redcap}{RedCap}{reduced capability}
\newacronym{phy}{PHY}{physical layer}
\newacronym{mac}{MAC}{medium access control}
\newacronym{kpi}{KPI}{key performance indicator}
\newacronym{ue}{UE}{user equipment}

\newacronym{otfs}{OTFS}{orthogonal time frequency space}
\newacronym{dd}{DD}{delay--Doppler}
\newacronym{cir}{CIR}{channel impulse response}
\newacronym{bler}{BLER}{block error rate}
\newacronym{nve}{NVE}{normalized validation error}
\newacronym{snr}{SNR}{signal-to-noise ratio}
\newacronym{llr}{LLR}{log-likelihood ratio}
\newacronym{ldpc}{LDPC}{low-density parity-check}
\newacronym{qam}{QAM}{quadrature amplitude modulation}
\newacronym{qpsk}{QPSK}{quadrature phase shift keying}
\newacronym{dft}{DFT}{discrete Fourier transform}
\newacronym{fft}{FFT}{fast Fourier transform}

\newacronym{ec}{EC}{expectation consistency}
\newacronym{ep}{EP}{expectation propagation}
\newacronym{mp}{MP}{message passing}
\newacronym{amp}{AMP}{approximate message passing}
\newacronym{uamp}{UAMP}{unitary approximate message passing}
\newacronym{lmmse}{LMMSE}{linear minimum mean square error}
\newacronym{mmse}{MMSE}{minimum mean square error}
\newacronym{pic}{PIC}{parallel interference cancellation}
\newacronym{sic}{SIC}{successive interference cancellation}

\newacronym{sdr}{SDR}{software-defined radio}
\newacronym{soc}{SoC}{system-on-chip}
\newacronym{dsp}{DSP}{digital signal processor}
\newacronym{vspa}{VSPA}{Vector Signal Processing Accelerator}
\newacronym{ippu}{IPPU}{Intervector Permutation Processing Unit}
\newacronym{dmem}{DMEM}{data memory}
\newacronym{pram}{PRAM}{program memory}
\newacronym{dma}{DMA}{direct memory access}
\newacronym{ram}{RAM}{random access memory}
\newacronym{isa}{ISA}{instruction set architecture}
\newacronym{iss}{ISS}{instruction set simulator}
\newacronym{pll}{PLL}{phase-locked loop}
\newacronym{rf}{RF}{radio frequency}
\newacronym{msps}{MSPS}{megasamples per second}
\newacronym{jtag}{JTAG}{Joint Test Action Group}
\newacronym{tap}{TAP}{test access port}
\newacronym{elf}{ELF}{executable and linkable format}
\newacronym{fpga}{FPGA}{field-programmable gate array}
\newacronym{gpu}{GPU}{graphics processing unit}
\newacronym{npu}{NPU}{neural processing unit}
\newacronym{pcie}{PCIe}{Peripheral Component Interconnect Express}
\newacronym{svd}{SVD}{singular value decomposition}

\newacronym{llm}{LLM}{large language model}
\newacronym{ai}{AI}{artificial intelligence}
\newacronym{dfm}{DFM}{design for manufacturability}
\newacronym{drc}{DRC}{design rule check}
\newacronym{hls}{HLS}{high-level synthesis}
\newacronym{rtl}{RTL}{register transfer level}
\newacronym{eda}{EDA}{electronic design automation}
\newacronym{oai}{OAI}{OpenAirInterface}
\newacronym{osa}{OSA}{OpenAirInterface Software Alliance}